\documentclass[a4paper,sn-mathphys]{sn-jnl}% Math and Physical Sciences Reference Style
\usepackage{fix-cm}
\usepackage{comment}
\usepackage{enumitem}
\usepackage{subcaption}

\begin{document}

\title[AI-based Learning Assistants]{Using AI-based Learning Assistants in Higher Education: A Large-Scale Descriptive Analysis}

\author{\fnm{Kristina} \sur{Schaaff}}\email{kristina.schaaff@iu.org}
\author{\fnm{Quintus} \sur{Stierstorfer}}\email{quintus.stierstorfer@iu.org}
\author{\fnm{Valerie} \sur{Hekkel}}\email{valerie.hekkel@iu.org}

\affil{\orgdiv{IU International University of Applied Sciences}, \orgaddress{\city{Erfurt}, \country{Germany}}}

%\documentclass[sn-mathphys]{sn-jnl}% Math and Physical Sciences Reference Style
%\usepackage{fix-cm}
%\usepackage{comment}
%\usepackage{enumitem}

%%%% Standard Packages
%%<additional latex packages if required can be included here>
%%%%
\abstract{In this study, we present a large-scale descriptive analysis of the use of an AI-based learning assistant (Syntea) in higher education. Based on objective log data from 77,543 students enrolled in distance studies, we examine usage patterns across gender, age group, study cluster, degree, and study mode. To date, existing research on educational chatbots has largely relied on comparatively small samples and self-reported survey data, while large-scale evidence on actual usage behavior remains limited. Our findings show that Syntea is already embedded in the study routines of many learners, but that usage differs across demographic and structural contexts. By identifying these patterns, our study provides an empirical basis for the further development of AI-based learning support and contributes a large-scale analysis of educational chatbot usage in higher education. }


\maketitle

\section{Introduction}

Educational chatbots and AI-based learning assistants are increasingly being implemented across various educational contexts, from primary to higher education, spanning diverse subjects from language learning \cite{YingSoon2024IntegratingAC, Lyu2024EffectivenessOC} to physics \cite{Bohomolova2021UsingCT, Lieb2024StudentIW} and nursing education \cite{chanIntegrationArtificialIntelligence2025}. Studies repeatedly highlight the potential of AI-based learning assistants to enhance student engagement \cite{Baskara2023ChatbotsAF}, provide personalized learning experiences \cite{HT2024HumanCE, mehlan2025}, and support administrative functions~\cite{SegoviaGarca2024OptimizingSS, ParralesBravo2024CSMAC}.

In our study, we investigate the usage of the AI-based learning assistant Syntea by students enrolled in distance studies at IU International University of Applied Sciences (IU). Students use Syntea via a chatbot interface to engage with their course content. We analyze usage patterns across key demographic factors, including gender, age, study cluster, degree, and study mode, to better understand how students interact with the system in their everyday learning practices. Accordingly, our study follows the research question of how the usage of AI-based learning assistants differs across demographic and structural student characteristics and across temporal study routines in distance education.

However, there remains a notable research gap regarding large-scale analysis of student demographic factors and their influence on interaction patterns with AI-based learning assistants, particularly in distance learning environments. While studies have examined specific implementations and outcomes, comprehensive analyses of usage patterns across demographics at an institutional level are limited. The current study on Syntea usage at IU addresses this gap by analyzing a large-scale dataset of interactions between students and Syntea across various demographic and structural dimensions.

This study makes three main contributions. First, it contributes large-scale descriptive evidence on the actual use of an AI-based learning assistant in higher education based on objective log data. Second, it identifies variation in usage patterns across demographic and structural factors, including gender, age, study cluster, degree, and study mode. Third, it provides an empirical foundation for the targeted refinement of AI-based learning support for diverse student populations in distance learning environments.

\section{Synthetic Teaching at International University of Applied Sciences}
Syntea is an AI-based learning assistant developed by IU to support students in their learning process and help them achieve their academic and career goals. It is integrated into the learning platform and also accessible via web and mobile applications, ensuring continuous availability within students’ existing learning environment \cite{möller2024revolutionisingdistancelearningcomparative}. 

Syntea supports a broad range of educational interactions, including factual question answering, open-ended chat, interactive tutoring, and personalized exam training. Additionally, it employs deep-dialog learning methods inspired by the Socratic approach, encouraging students to reflect and engage critically with the material. The system can also provide personalized feedback, track learning progress, and adapt to individual needs, thereby supporting self-paced learning.
It is capable of personalizing learning content based on the students' career goals, which has been shown to increase reflection depth, identification with the learning content, and efficiency of learning trajectories~\cite{mehlan2025}.

From a technological perspective, Syntea is built on state-of-the-art generative AI models. During the test period, those were GPT-4, GPT-4-Turbo, and GPT-3.5-Turbo.

\section{Usage Patterns of AI-based Chatbots in Education}

AI-based chatbots have become important tools in education \cite{labadzeRoleAIChatbots2023}. 
A study of UAE universities found that most of 210 undergraduate students involved in the study were willing to transition from physical to online educational platforms \cite{Ali2021TheST}.

Stöhr et al. \cite{STOHR2024100259} reported significant differences in AI-based chatbot use and perceptions across gender, academic level, and field of study, based on survey data from 5,894 students in Sweden. Male students used chatbots more regularly and showed more positive attitudes, while female students expressed more skepticism. Master’s students reported higher usage than undergraduates, and students in technology and engineering showed the highest usage rates. In a study of 363 students at a public university in the United States, Deng et al. \cite{Deng2025Demographics} found higher adoption for generative AI among students in advanced academic years, non-native English speakers, and students of Asian descent. Both studies rely on self-reported survey data and lack objective measures of actual chatbot use. Moreover, the sample in \cite{STOHR2024100259} includes students from various universities, which may limit generalizability.

Gezgin et al. \cite{gezgin2024} studied LLM-based chatbot use in programming education with 413 students and found significant differences by gender and grade level, but not by department. Pazzaglia et al. \cite{pazzaglia2016engagement} analyzed behavioral engagement in online high school courses using log data from 5,511 enrollments, but did not focus on chatbot use or student characteristics, and was conducted prior to the emergence of LLM-based systems.

\begin{table}[ht]
  \centering \tiny
  \caption{Sample Sizes and Data Types in Related Chatbot and Online Learning Studies.}
  \label{tab:related-work-comparison}
  \begin{tabular}{llrl}
  \toprule
  \textbf{Study} & \textbf{Context} & \textbf{N} & \textbf{Data Type} \\
  \midrule
  Ali~\cite{Ali2021TheST}           & UAE universities        & 210   & Survey \\
  Deng et al.~\cite{Deng2025Demographics} & US universities & 363 & Survey\\
  Gezgin et al.~\cite{gezgin2024} & Programming education & 413   & Survey \\
  Pazzaglia et al.~\cite{pazzaglia2016engagement} & Online high school  & 5,511 & Log data \\
  Stöhr et al.~\cite{STOHR2024100259}  & Swedish universities   & 5,894 & Survey \\
  \textbf{Our study}           & \textbf{Distance university} & \textbf{77,543} & \textbf{Log data} \\
  \bottomrule
  \end{tabular}
  \label{tab:studycomparison}
  \end{table} 
  
Existing studies on educational chatbot usage rely on comparatively small samples and predominantly self-reported survey data (Table~\ref{tab:related-work-comparison}). In contrast, our study addresses both limitations by analyzing objective log data from 77,543 enrolled students and, to the best of our knowledge, provides the largest empirical analysis of AI teaching assistant usage patterns in higher education to date.
  
\section{Methodology}

This chapter describes the methodological approach, including data selection, cleaning procedures, and sample characteristics underlying the analysis of Syntea usage.

\subsection{Data Selection}

We collected statistical data from active online students about their Syntea usage behavior in February 2025. Metrics analyzed include usage distributions by gender and age groups, study cluster (a family of thematically related degree programs), degree, and study mode (i.e., full time or part time). 
February 2025 was selected to ensure balanced weekday frequencies and reduce seasonal distortions. Restricting the analysis to a single month also supports comparability and consistent logging conditions. Moreover, a phased migration to a new learning system started in March 2025. Therefore, in February 2025, usage patterns are not distorted by migration-related effects such as parallel-system use, shifting routines, or changes in tracking and instrumentation that typically become more pronounced as adoption increases. 

Restricting the observation period to a single month further limits heterogeneity that can arise from curricular changes. A narrowly defined window also increases the likelihood of consistent logging quality (e.g., stable event definitions and uniform tracking procedures), thereby supporting internal validity and simplifying preprocessing. 

\subsection{Data Preparation and Cleaning}

For our study, we included all students enrolled in Bachelor's or Master's programs in distance studies in the month of February 2025. We only included students who started the program before Februrary 01, 2025, and finished after Februrary 28, 2025. 
Additionally, students who were enrolled in more than one degree program were excluded from the analysis, as their performance could not always be unambiguously assigned to a single program.

During data cleaning, 78 incomplete records (e.g., originating from test users) were removed from the dataset.

For the analysis of age-related differences, we classified participants into generational cohorts following the definitions of \cite{Pew2019Generations}. The corresponding birth year ranges, and the labels used in this paper are summarized in Table~\ref{tab:generations_pew}. At the time of our analyses, there were no students enrolled out of these ranges. 

\begin{table}[t]
\centering\tiny
\caption{Generational Cohort Definitions.}
\label{tab:generations_pew}
\begin{tabular}{llr}
\toprule
\textbf{Generation} & \textbf{Label} & \textbf{Birth Years} \\
\midrule
Baby Boomers & Boomers & 1946--1964 \\
Generation X & Gen X & 1965--1980 \\
Millennials / Generation Y  & Gen Y & 1981--1996 \\
Generation Z & Gen Z & 1997--2012 \\
\bottomrule
\end{tabular}
\end{table}

\subsection{Data Description}

After data cleaning, our analysis included 76,485 students enrolled in distance studies at IU, as at the time of our research, Syntea was only offered in distance learning. Table \ref{tab:characteristics} provides an overview of the student characteristics. 

Overall, the sample is predominantly female (64.72\%) and largely enrolled in Bachelor’s programs (84.77\%). Enrollments are distributed across study clusters with the highest shares in Health \& Social (22.41\%), Business \& Management (20.99\%), and Education \& Psychology (20.19\%). Participation is nearly evenly split between part-time (50.78\%) and full-time (49.22\%) study modes. With respect to age, most students belong to Gen~Z (51.44\%) and Gen~Y (42.72\%), indicating a comparatively young student population.

\begin{table}[ht] %2026-02-22
\tiny
\centering
\caption{Descriptive Characteristics of the Study Participants}
\begin{tabular}{@{}lrr@{}}
\toprule
\textbf{Characteristics} & \textbf{Frequency} & \textbf{Proportion (\%)} \\
\midrule
\textbf{Gender} & & \\
female & 49,500 & 64.72 \\
male & 26,876 & 35.14 \\
diverse & 109 & 0.14 \\
\midrule
\textbf{Age Group} & & \\
Gen Z & 39,343 & 51.44 \\
Gen Y & 32,675 & 42.72 \\
Gen X & 4,335 & 5.67 \\
Boomers & 132 & 0.17 \\
\midrule
\textbf{Study Cluster} & & \\
Health \& Social & 17,138 & 22.41 \\
Business \& Management & 16,053 & 20.99 \\
Education \& Psychology & 15,440 & 20.19 \\
IT \& Technology & 11,705 & 15.30 \\
Marketing \& Communication & 4,574 & 5.98 \\
Human Resources \& Law & 3,859 & 5.05 \\
Architecture \& Construction & 3,298 & 4.31 \\
Design \& Media & 3,023 & 3.95 \\
Tourism \& Hospitality & 1,395 & 1.82 \\
\midrule
\textbf{Degree} & & \\
Bachelor & 64,836 & 84.77 \\
Master & 11,649 & 15.23 \\
\midrule
\textbf{Study Mode} & & \\
Part Time & 38,837 & 50.78 \\
Full Time & 37,648 & 49.22 \\

\bottomrule
\end{tabular}
\label{tab:characteristics}
\end{table}

\section{Analysis of Syntea Usage}

This chapter presents the descriptive results of Syntea usage, focusing on overall adoption patterns and differences across key student characteristics.

\subsection{Overall Syntea Usage}

During our observation period, a total of 44,035 out of the 76,485 students in our sample used Syntea. 2,509 students used Syntea for the first time during this period, while 41,526 students were returning users.

Overall, this indicates that Syntea was already used by a substantial proportion of students during the study period. There are several reasons why not all students used Syntea. For instance, Syntea is not available in all courses, especially project courses or seminars. Moreover, students who begin their final thesis may stop using Syntea. Temporary study interruptions, e.g., due to vacation or illness, can also lead to periods in which Syntea is not used. However, some students may have disliked Syntea or may be opposed to AI tools.

\subsection{Analysis by Gender}

Table \ref{tab:counts_gender} shows the number of students using Syntea by gender. 
\begin{table}[ht]
\tiny
\centering
\caption{Counts by Gender (Total and Syntea users)}
\begin{tabular}{@{}lrrr@{}}
\toprule
\textbf{Gender} & \textbf{Total} & \textbf{Syntea Users} & \textbf{Percentage} \\
\midrule
female & 49,500 & 29,230 & 59.05\\
male & 26,876 & 14,767 & 54.94\\
diverse & 109 & 38 & 34.86\\
\bottomrule
\end{tabular}
\label{tab:counts_gender}
\end{table}
The percentage of \textit{female} students (59.05\%) using Syntea during our observation period is slightly higher than the percentage of \textit{male} students (54.94\%). Only 34.86\% of the \textit{diverse} students were using Syntea during the observation period. However, due to the small sample size, this percentage has limited explanatory power.
These differences between \textit{males} and \textit{females} may reflect unequal exposure across programs: Social science disciplines, which often enroll a higher proportion of \textit{female} students, integrate Syntea into a larger number of courses, thereby increasing opportunities for use.

\subsection{Analysis by Age Group}
\label{sec:analysis-age}             

The age distribution differs slightly between Syntea users and non-users. Non-users had a mean age of 30.44 years (SD = 7.52), with ages ranging from 18 to 71 years (IQR: 25–34; median = 28). Syntea users (n = 44,035) were younger on average (M = 28.69, SD = 7.13), with ages ranging from 16 to 75 years (IQR: 24–32; median = 27).

\begin{figure}[htp]
    \centering
    \includegraphics[scale=0.3]{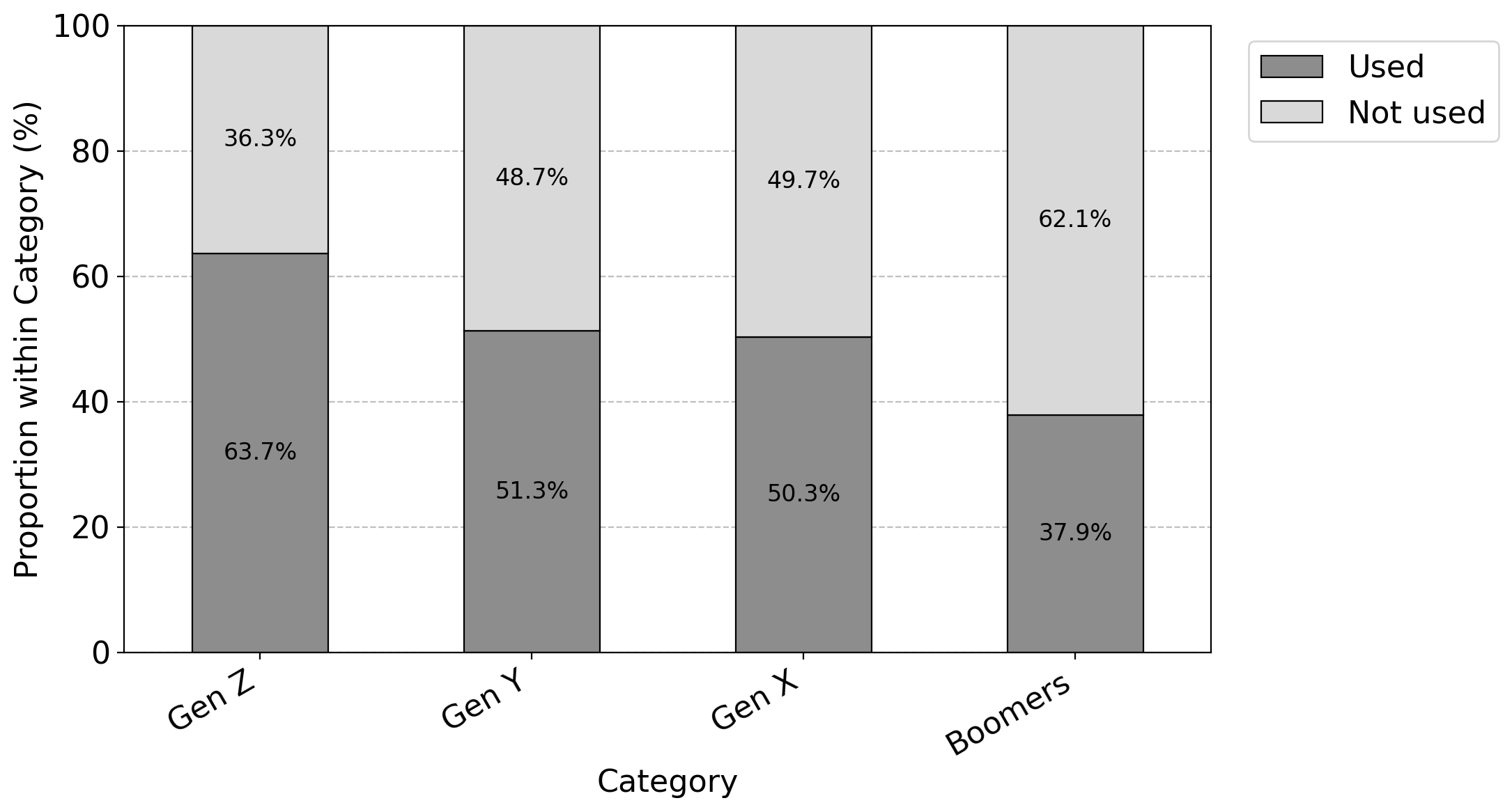}
    \caption{Percentage of Syntea Users by Age Group}
    \label{fig:syn_age}
\end{figure}

Figure~\ref{fig:syn_age} shows Syntea usage rates by generational cohort during the observation period. \textit{Gen~Z} students exhibited the highest usage rate at 63.66\%, followed by \textit{Gen~Y} with 51.39\% and \textit{Gen~X} with 50.27\%. \textit{Boomers} showed a usage rate of approximately 37.88\%; however, given the very small cohort size ($n = 132$), this estimate is subject to considerable sampling variability and should be interpreted with caution.

The notably higher adoption among \textit{Gen~Z} students may reflect a greater familiarity and comfort with AI-based interfaces, consistent with findings from Stöhr et al.~\cite{STOHR2024100259}, who observed that younger students reported more positive attitudes toward AI chatbots in educational settings. However, this cohort effect may also be partially confounded by study phase: \textit{Gen~Z} students are more likely to be in the earlier semesters of their Bachelor's programs, where Syntea coverage is highest, and thesis work (where Syntea is unavailable) has not yet begun. A similar confound may apply to the \textit{Boomer} cohort, where the still high usage rate could reflect a self-selection effect: the few students in this age group who enroll in distance studies may be particularly motivated and technology-open, rather than representative of their generation at large.

\subsection{Analysis by Study Cluster}
\begin{figure}
    \centering
    \includegraphics[scale=0.29]{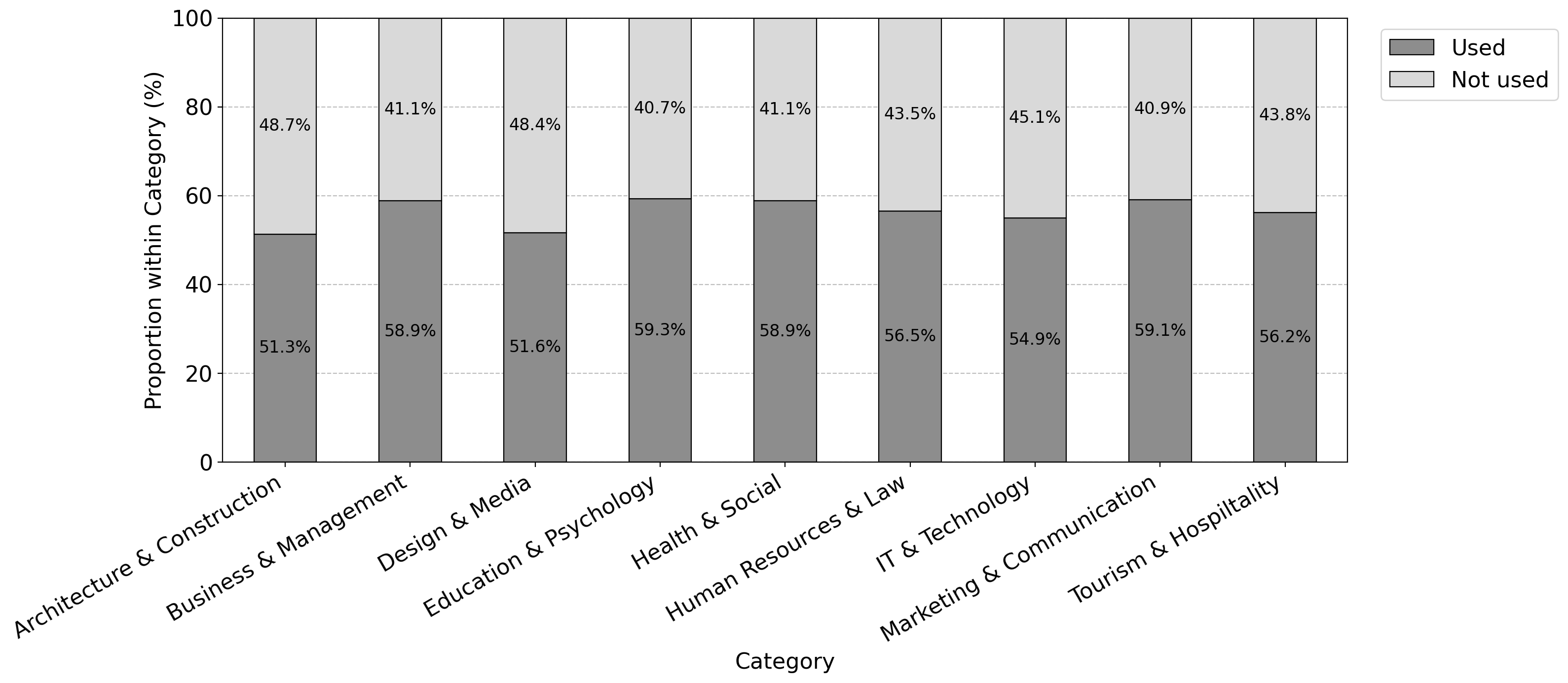}
    \caption{Percentage of Syntea Users by Study Cluster}
    \label{fig:syn_study_cluster}
\end{figure}

Figure \ref{fig:syn_study_cluster} illustrates how Syntea is used in different study clusters. In all categories, the usage rate is relatively close. It is highest in \textit{Education \& Psychology} and \textit{Marketing \& Communication}, and lowest in \textit{Architecture \& Construction} and \textit{Design \& Media}. 

The comparatively small differences across study clusters suggest that Syntea is broadly applicable across disciplines. However, the lower usage in more visually oriented fields such as \textit{Architecture \& Construction} or \textit{Design \& Media} indicates that the current implementation may be better aligned with text-based and concept-heavy learning activities than with disciplines that depend more strongly on visual, practical, or software-based work.

\subsection{Analysis by Degree}

Table \ref{tab:counts_degree} shows that Syntea usage was slightly higher among \textit{Bachelor}’s students than among \textit{Master}’s students. At first glance, this may suggest lower adoption in \textit{Master}’s programs. However, this difference should be interpreted with caution, as \textit{Master}’s programs are typically shorter in duration and therefore include a proportionally larger thesis phase, during which Syntea is generally not used. Consequently, the observed difference may partly reflect structural differences in program composition rather than lower acceptance or demand among \textit{Master}’s students.

\begin{table}[ht]
\tiny
\centering
\caption{Counts by Degree (Total and Syntea Users)}
\begin{tabular}{@{}lrrr@{}}
\toprule
\textbf{Degree} & \textbf{Total} & \textbf{Syntea Users} & \textbf{Percentage} \\
\midrule
Bachelor & 64,836 & 37,715 & 58.17\\
Master & 11,649 & 6,320 & 54.25\\
\bottomrule
\end{tabular}
\label{tab:counts_degree}
\end{table}

\subsection{Analysis by Study Mode}

Table \ref{tab:counts_mode} shows only small differences in Syntea usage by study mode. During the observation period, 59.49\% of \textit{full-time} students used Syntea, compared with 55.71\% of \textit{part-time} students. This slightly higher adoption among \textit{full-time} students may reflect structural differences in study organization: \textit{full-time} students may engage more continuously with course content, whereas \textit{part-time} students often distribute their learning activities more flexibly around work or other obligations. Accordingly, the difference should not be interpreted as a difference in acceptance alone, but may also reflect variation in available study time and usage opportunities.

\begin{table}[ht]
\tiny
\centering
\caption{Counts by Study Mode (Total and Syntea Users)}
\begin{tabular}{@{}lrrr@{}}
\toprule
\textbf{Study Mode} & \textbf{Total} & \textbf{Syntea Users} & \textbf{Percentage} \\
\midrule
Part Time & 38,837 & 21,637 & 55.71\\
Full Time & 37,648 & 22,398 & 59.49\\
\bottomrule
\end{tabular}
\label{tab:counts_mode}
\end{table}

\section{Temporal Usage Patterns}

In this section, we will examine students’ Syntea usage patterns over time, based on the time of the day and the weekday.

Temporal usage patterns are derived from hourly activity counts. Activity is counted once per unique combination of student, hour, and day (of the month), i.e., a value of 1 is added to the count if a student was active during that hour and day, and 0 otherwise.
This definition applies both to hour-based and weekday-based aggregations. Across all students, usage was normalized across hours of the day and days of the week such that the respective totals sum to 100\% within each user group.

\subsection{Overall Analysis of Temporal Usage Patterns}

\begin{figure}[htbp]
    \centering

    \begin{subfigure}[t]{0.7\textwidth}
        \centering
        \includegraphics[width=\textwidth]{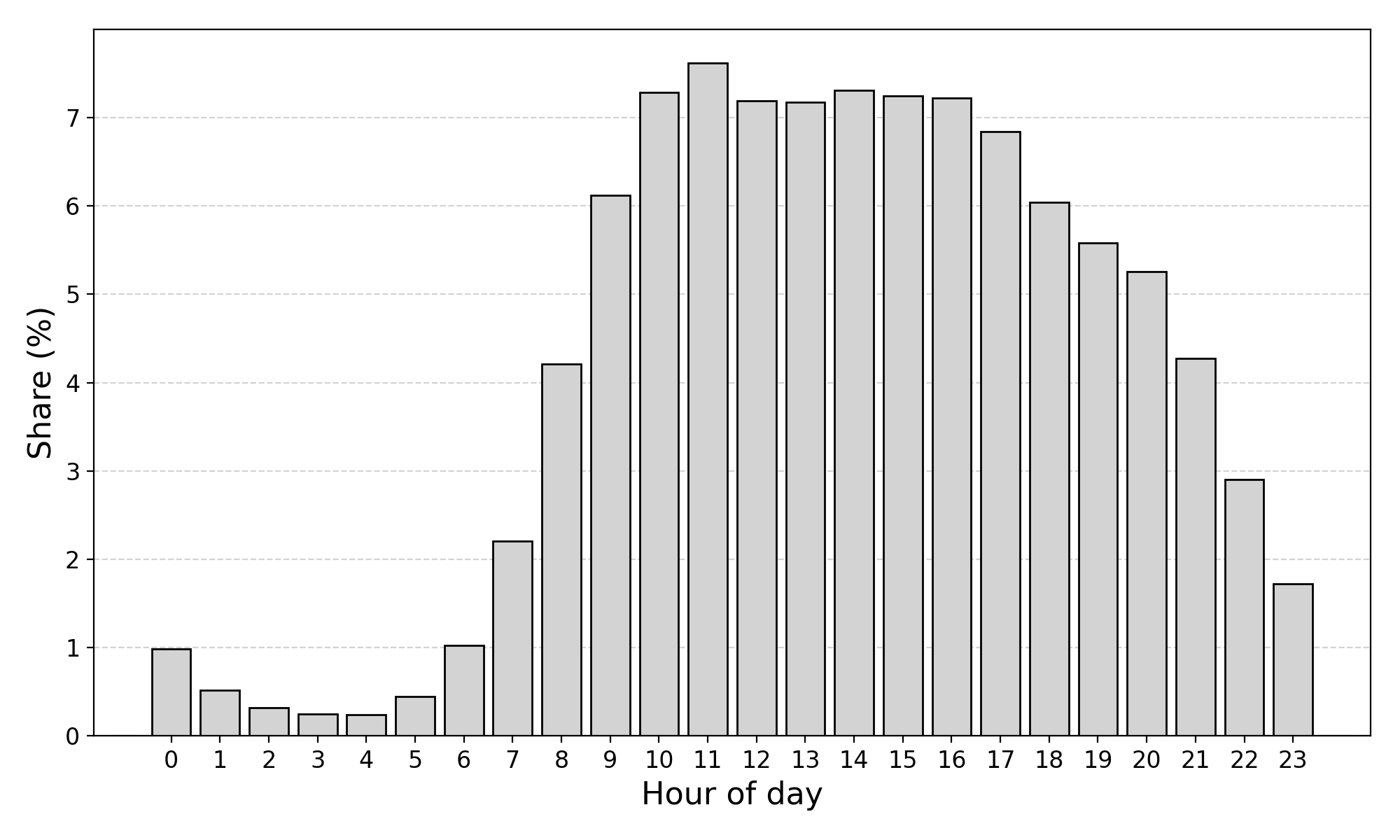}
       % \label{fig:usage_hour_all}
    \end{subfigure}
    \hfill
    \begin{subfigure}[t]{0.28\textwidth}
        \centering
        \includegraphics[width=\textwidth]{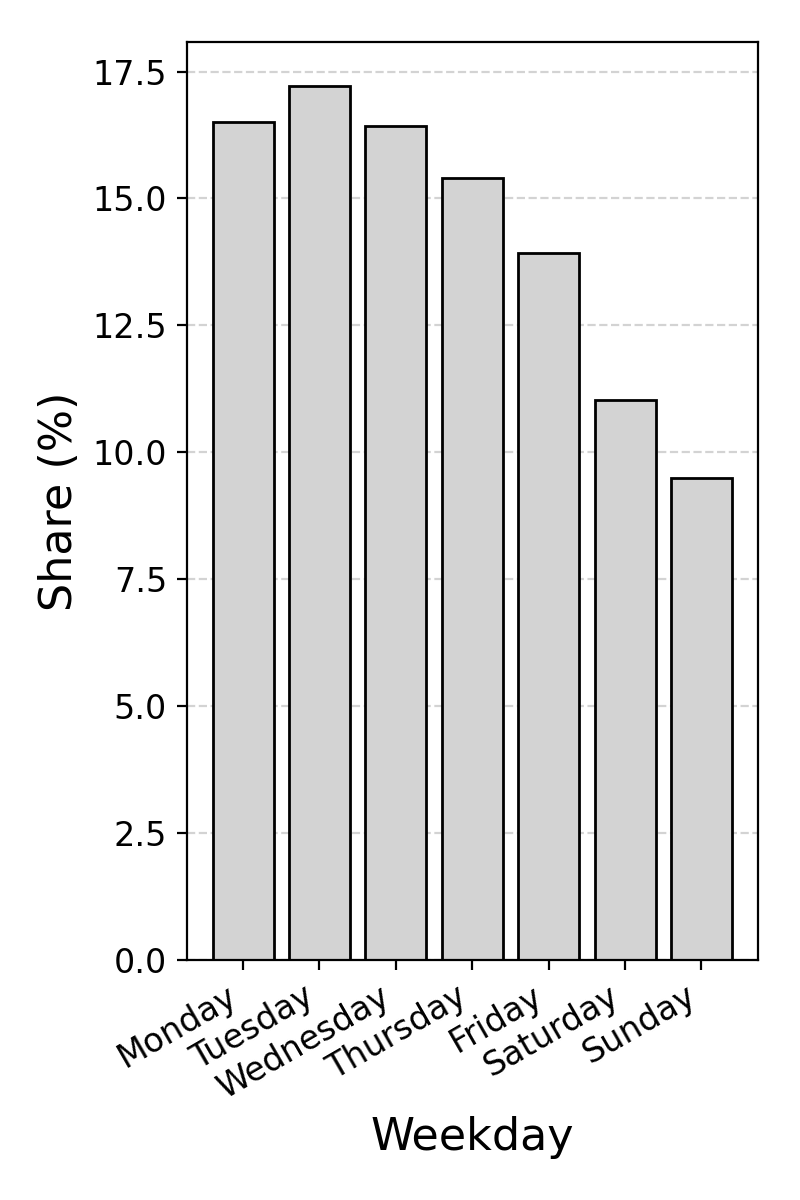}
       % \label{fig:weekday_overall_ALL}
    \end{subfigure}

    \caption{Normalized overall usage patterns for all students by hour (left) and weekday (right)}
    \label{fig:overall_usage_all}
\end{figure}

Figure \ref{fig:overall_usage_all} illustrates the temporal usage patterns of Syntea among students across the day and the week. 

The hourly distribution shows that usage is very low during the night and early morning hours, with the smallest shares occurring between approximately 2:00 and 5:00. Activity begins to increase noticeably from around 7:00, rises sharply between 8:00 and 10:00, and reaches its highest level in the late morning, peaking around 11:00. Usage remains consistently high throughout the early and mid-afternoon hours, particularly between 10:00 and 16:00, before gradually declining from the late afternoon onward. By the late evening, usage decreases substantially again.

The weekday distribution indicates that Syntea is used most intensively on weekdays and less frequently on weekends. The highest share is observed on Tuesday, followed closely by Monday and Wednesday. Usage declines slightly on Thursday and Friday, and drops more markedly on Saturday and Sunday, with Sunday showing the lowest overall share. Taken together, these findings suggest that students primarily use Syntea during regular daytime study hours and predominantly on working days, indicating that usage is closely aligned with typical academic routines.

\subsection{Analysis by Gender}

\begin{figure}
    \centering
    \includegraphics[scale=0.3]{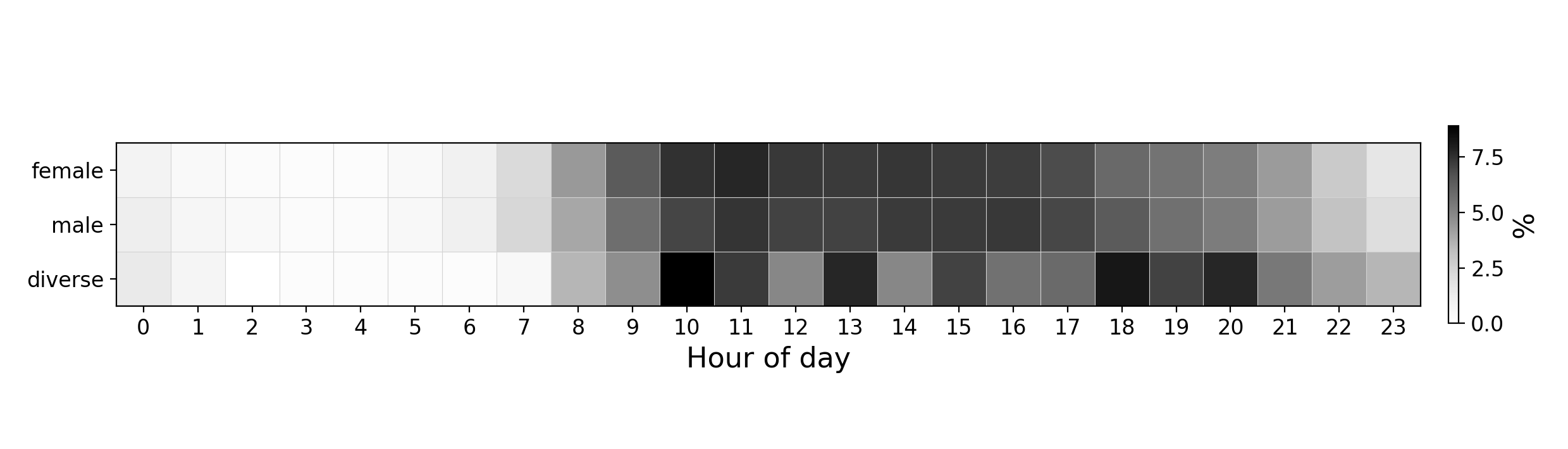}
    \caption{Usage Hours by Gender}
    \label{fig:usage_hour_gender}
\end{figure}

\begin{figure}
    \centering
    \includegraphics[scale=0.3]{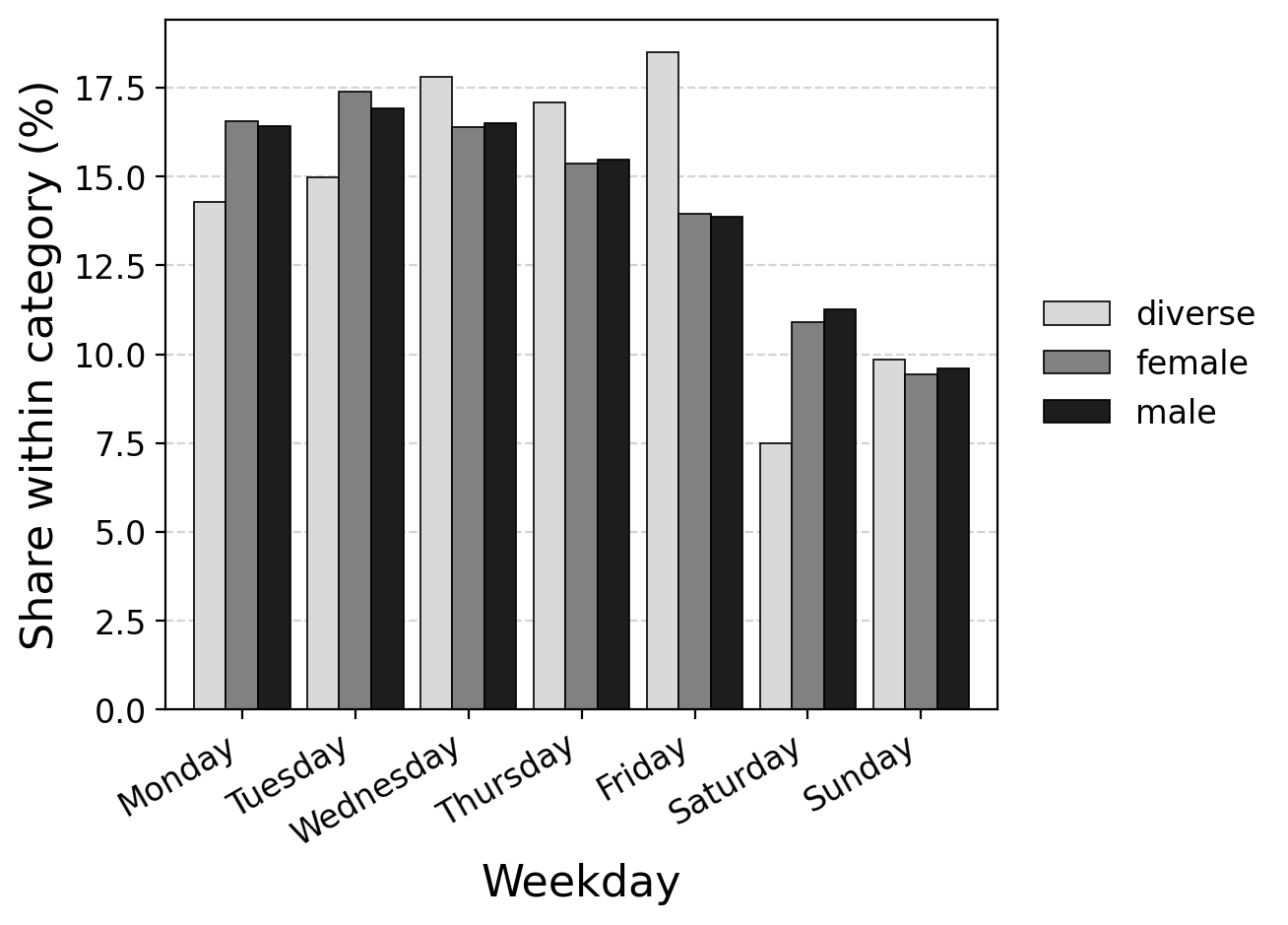}
    \caption{Syntea Usage by Weekday by Gender}
    \label{fig:weekday_gender}
\end{figure}

The temporal usage patterns of Syntea were broadly similar across gender groups. Across all genders, activity was concentrated during daytime to early evening hours (Figure \ref{fig:usage_hour_gender}) and on weekdays (Figure \ref{fig:weekday_gender}), indicating that Syntea use largely followed regular study routines rather than late-night learning behavior. With regard to hourly usage, \textit{female} and \textit{male} students showed highly comparable patterns: activity was minimal during the night, increased markedly in the morning, and reached its highest levels between late morning and the afternoon, before gradually declining in the evening. Thus, gender differences appear to be limited more to overall adoption rates than to fundamentally different temporal usage structures.

The weekday distribution also showed a largely consistent pattern across \textit{female} and \textit{male} students. In both groups, usage was highest on weekdays—particularly early in the week—and declined toward the weekend, with the lowest shares observed on Sunday. This suggests that Syntea use is closely aligned with structured study schedules across genders. Although the descriptive plots suggest somewhat more irregular weekday and hourly patterns for students recorded as \textit{diverse}, these results should be interpreted with caution due to the very small size of this group.

\subsection{Analysis by Age Group}

\begin{figure}
    \centering
    \includegraphics[scale=0.3]{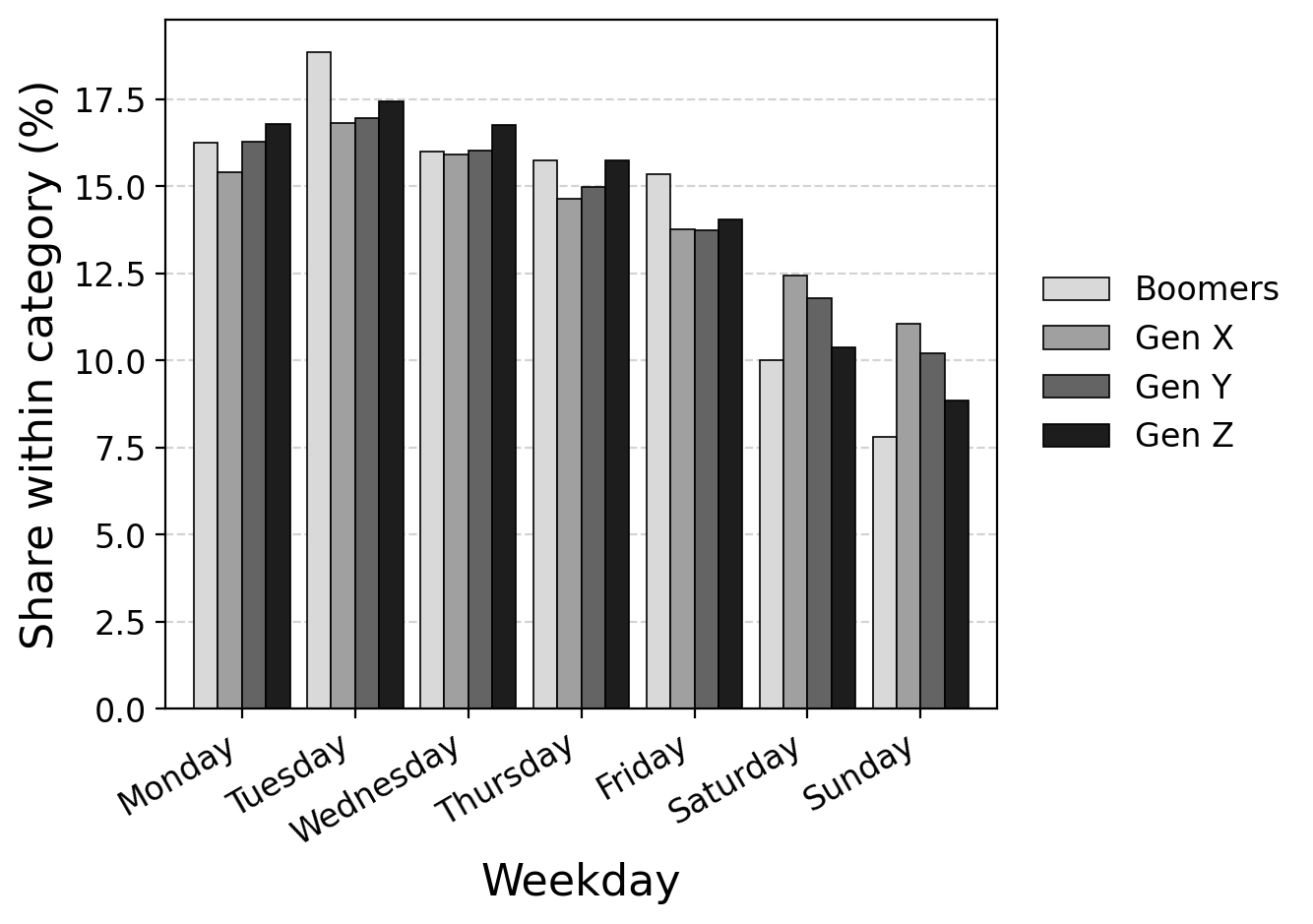}
    \caption{Syntea Usage by Weekday per Age Group}
    \label{fig:weekday_agegroup}
\end{figure}

\begin{figure}
    \centering
    \includegraphics[scale=0.3]{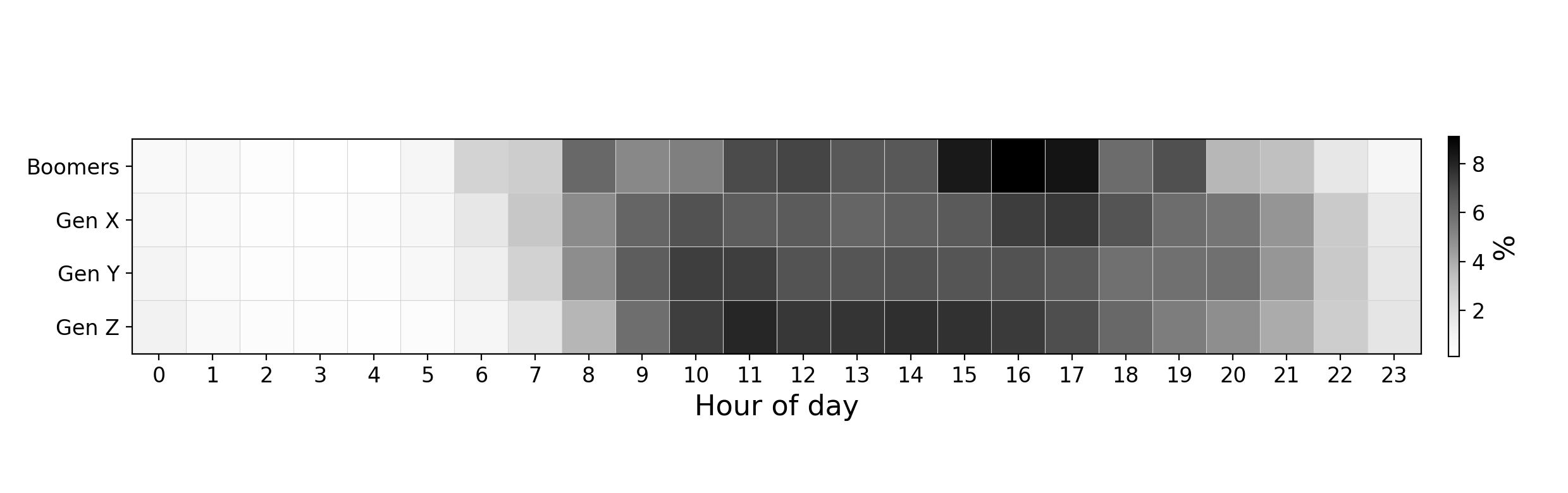}
    \caption{Syntea Usage by Hour per Age Group}
    \label{fig:usage_hour_age_group}
\end{figure}

Temporal usage patterns differed notably across age groups. Figure \ref{fig:weekday_agegroup} shows that all age groups used Syntea more on weekdays than on weekends, but the distribution across the week varied. \textit{Gen~Z} students showed the strongest concentration at the beginning of the week and a clearer decline toward the weekend, whereas \textit{Gen~X} and \textit{Gen~Y} were distributed somewhat more evenly. In the hourly analysis (Figure \ref{fig:usage_hour_age_group}), all age groups used Syntea primarily during daytime hours, but their daily profiles were not identical. \textit{Gen~Z} usage was more strongly concentrated from late morning to early afternoon and remained visible into the evening, while \textit{Gen~X} and \textit{Gen~Y} showed broader distributions across the day. \textit{Boomers} appeared to exhibit a later and more pronounced afternoon peak; however, this finding should be interpreted with caution due to the very small size of this group. Overall, the results point to shared general usage tendencies but clearly different temporal profiles across age groups.

\subsection{Analysis by Study Cluster}

\begin{figure}[h!]
    \centering
    \includegraphics[scale=0.3]{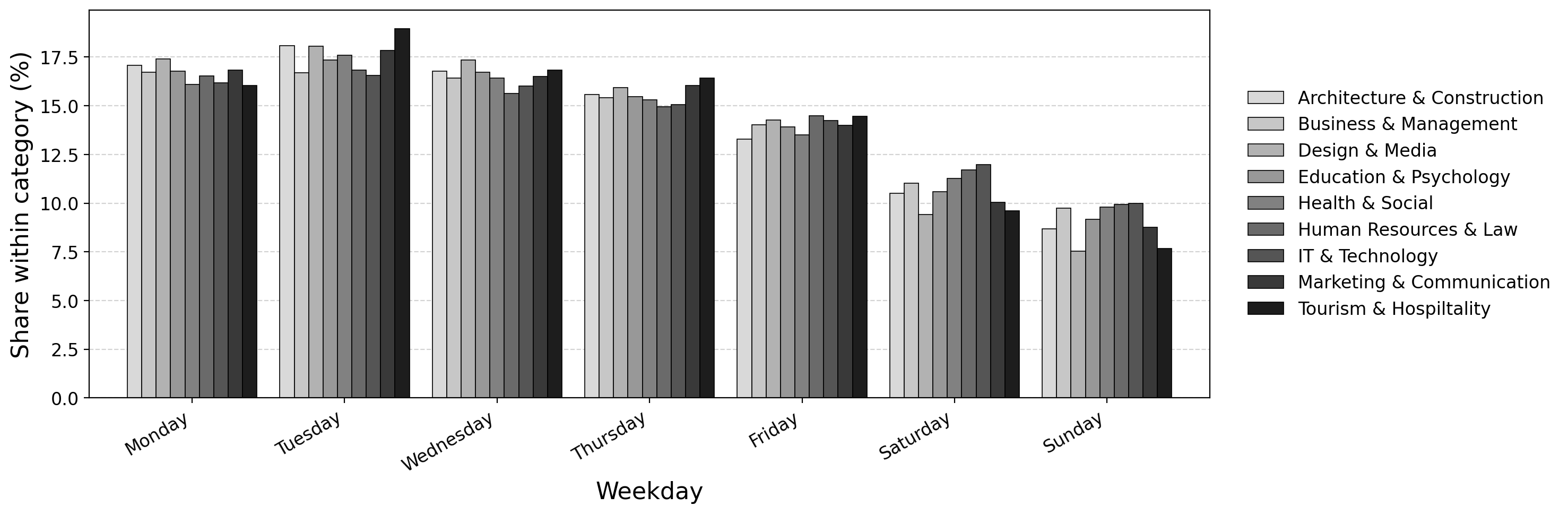}
    \caption{Syntea Usage by Weekday per Study Cluster}
    \label{fig:weekday_study_cluster}
\end{figure}
Figure \ref{fig:weekday_study_cluster} illustrates Syntea usage by weekday across study clusters. Although inter-cluster differences are generally small, some variations can be identified, especially on weekends. Clusters, such as \textit{Human Resources \& Law} or \textit{Business \& Management}, maintain slightly higher usage levels on Saturday, while others, such as \textit{Tourism \& Hospitality}, show lower shares, particularly on Sunday. Overall, however, the similarities between study clusters clearly outweigh the differences, suggesting that Syntea usage is primarily driven by general weekly study routines rather than discipline-specific patterns.

\begin{figure}[htp]
    \centering
    \includegraphics[scale=0.3]{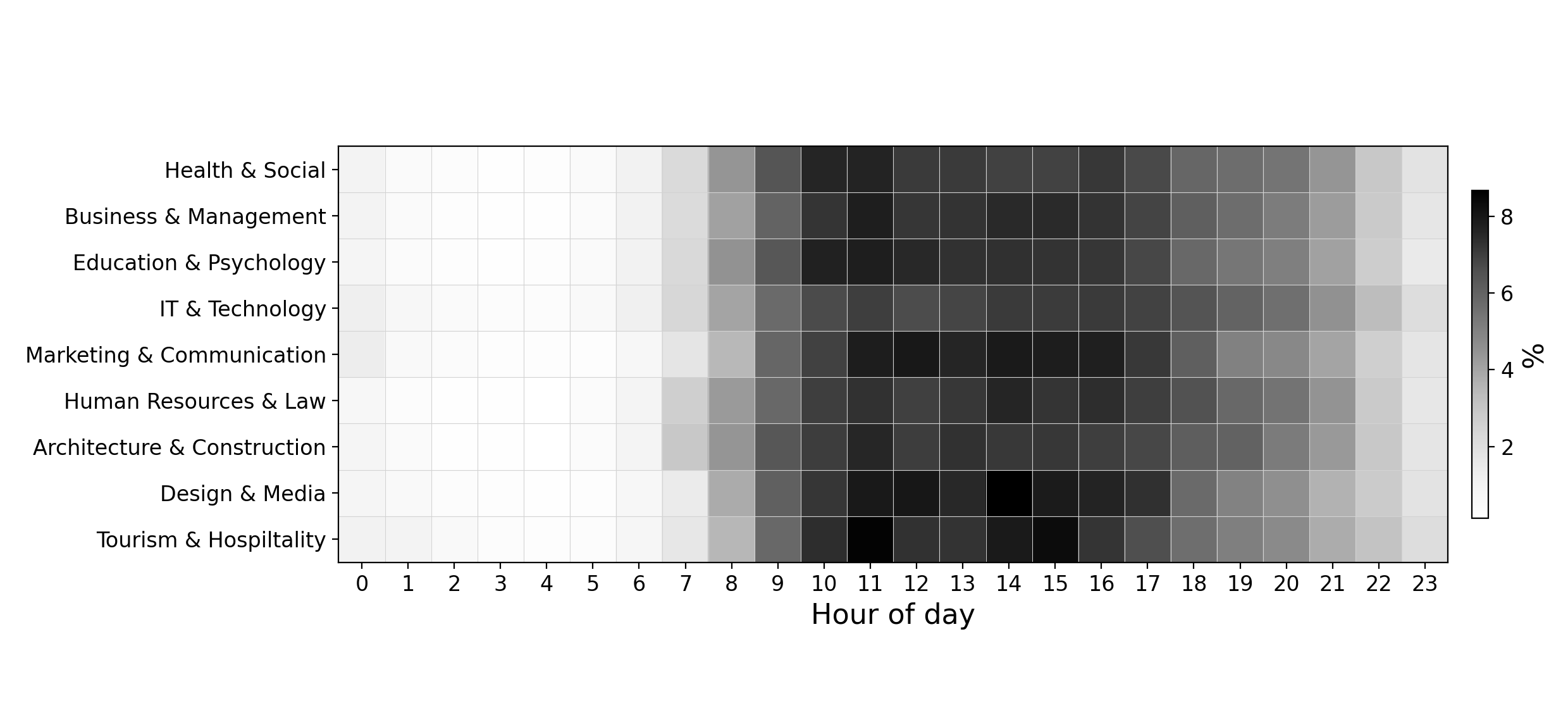}
    \caption{Syntea Usage by Hour per Study Cluster}
    \label{fig:usage_hour_study_cluster}
\end{figure}

The heatmap in Figure \ref{fig:usage_hour_study_cluster} indicates a highly consistent diurnal usage pattern across all study clusters. 
Differences between study clusters are mainly gradual rather than structural. \textit{Marketing \& Communication} and \textit{Design \& Media} exhibit the strongest concentration during the mid-day to afternoon period (particularly around 13:00–15:00). \textit{Tourism \& Hospitality} shows a similarly pronounced mid-day peak (approximately 11:00–15:00). In contrast, \textit{IT \& Technology} appears slightly more evenly distributed across the day, with comparatively sustained activity into the early evening. Overall, the pattern suggests predominantly study- or work-related use aligned with typical daytime schedules.

\subsection{Analysis by Degree}

\begin{figure}[htp]
    \centering
    \includegraphics[scale=0.3]{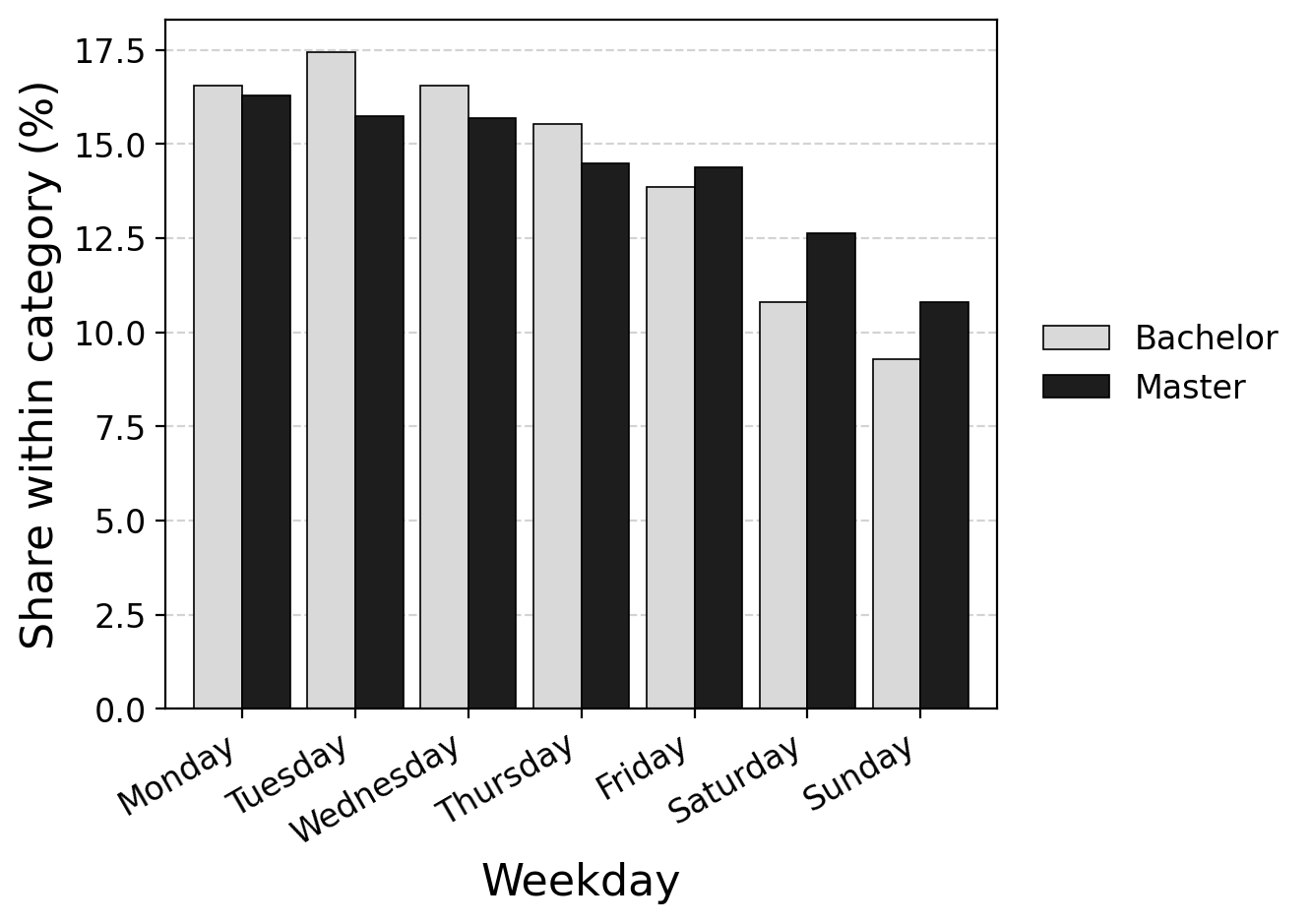}
    \caption{Usage by Weekday per Degree}
    \label{fig:weekday_degree}
\end{figure}

\begin{figure}[htp]
    \centering
    \includegraphics[scale=0.3]{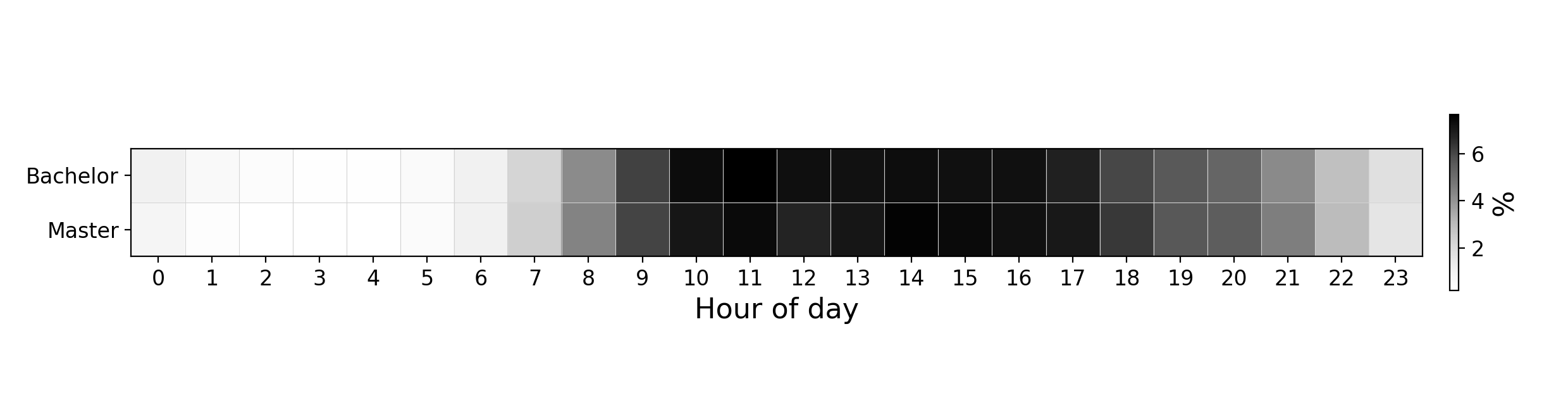}
    \caption{Syntea Usage by Hour per Degree}
    \label{fig:usage_hour_degree}
\end{figure}

As Figure \ref{fig:weekday_degree} and \ref{fig:usage_hour_degree} show, temporal usage patterns differ slightly by degree. In the weekday analysis, \textit{Bachelor} students showed a stronger concentration of Syntea usage during the early part of the week, with the highest shares on Tuesday and Wednesday, followed by a steady decline toward the weekend. \textit{Master} students, by contrast, displayed a slightly flatter weekday distribution, with comparatively lower shares at the beginning of the week but somewhat higher activity on Friday, Saturday, and Sunday. 

The hourly analysis indicates that both groups used Syntea primarily during daytime and early evening hours, but with slight differences in timing. \textit{Bachelor} students showed slightly stronger usage than \textit{Master} students in the late morning and around noon, whereas \textit{Master} students exhibited relatively higher shares from the early afternoon into the evening. 

\subsection{Analysis by Study Mode}

\begin{figure}
    \centering
    \includegraphics[scale=0.3]{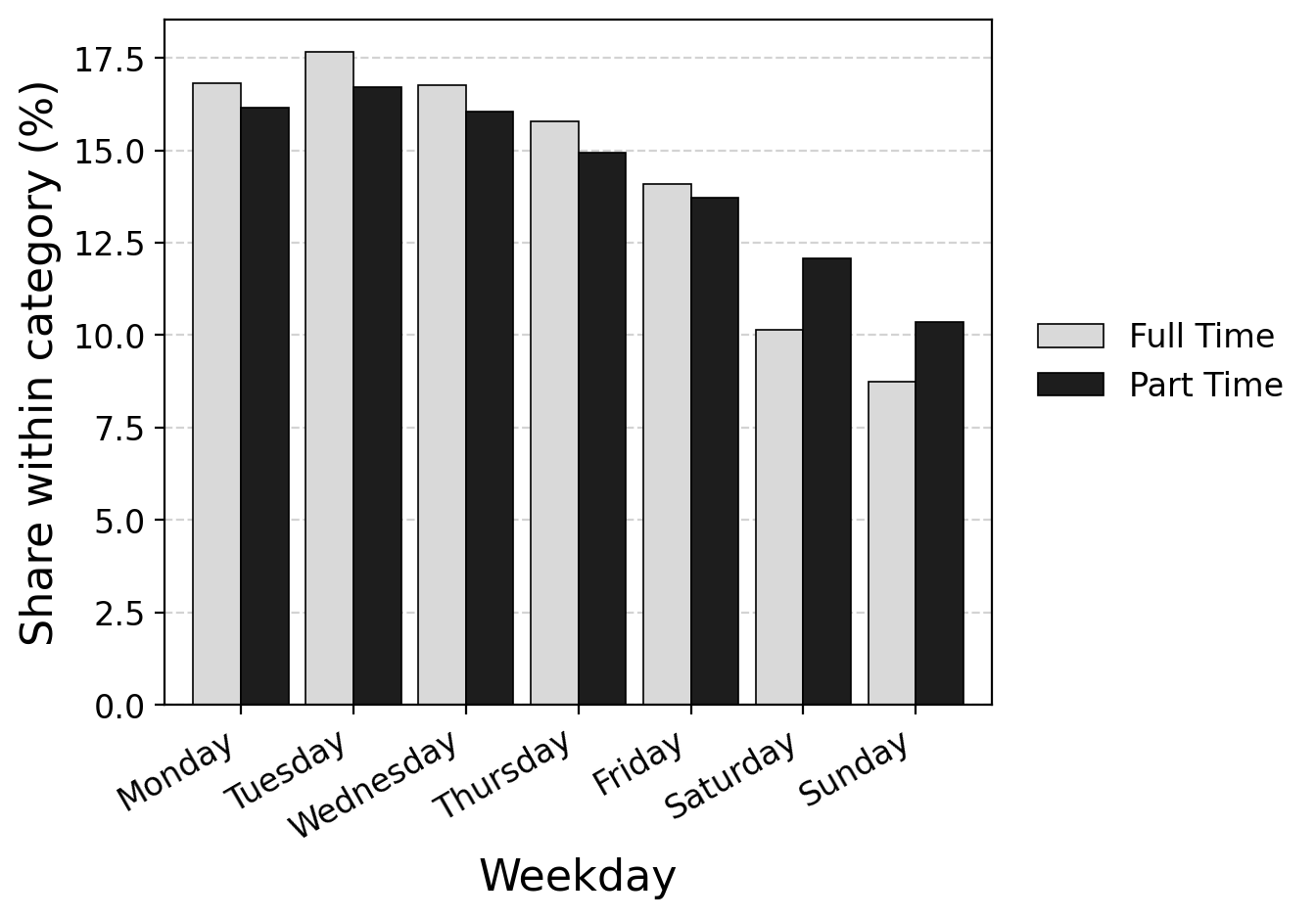}
    \caption{Syntea Usage by Weekday per Study Mode}
    \label{fig:weekday_studymode}
\end{figure}

\begin{figure}
    \centering
    \includegraphics[scale=0.3]{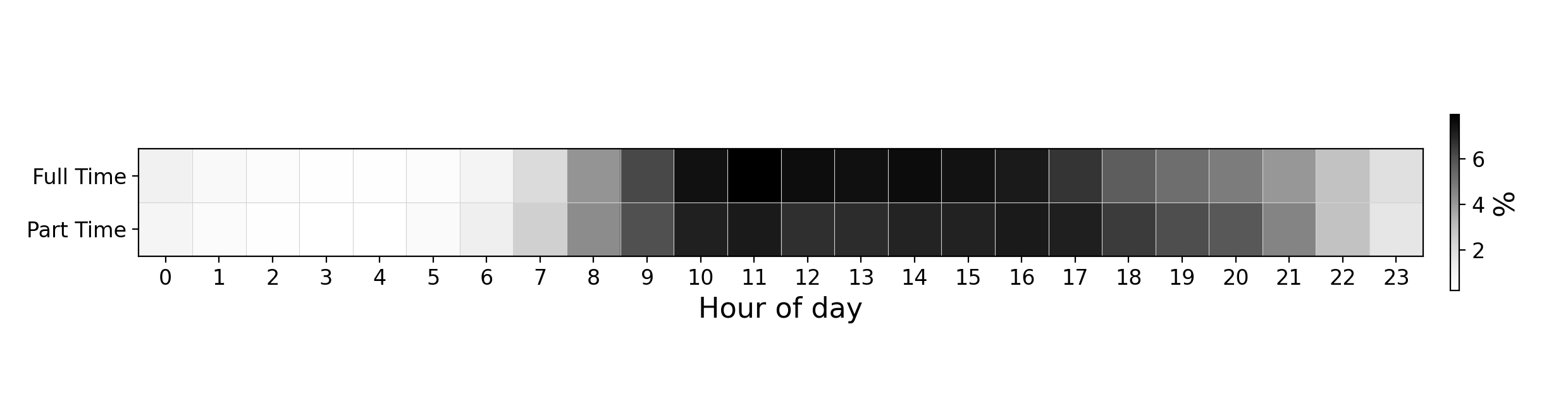}
    \caption{Syntea Usage by Hour per Study Mode}
    \label{fig:usage_hour_study_mode}
\end{figure}

Temporal usage patterns also varied by study mode. \textit{Full-time} students showed a stronger concentration of Syntea use at the beginning of the week (Figure~\ref{fig:weekday_studymode}) and during late morning to afternoon hours (Figure~\ref{fig:usage_hour_study_mode}). \textit{Part-time} students, by contrast, exhibited a somewhat flatter weekday pattern and relatively stronger usage on weekends as well as in the late afternoon and evening. These differences suggest that Syntea use among \textit{part-time} students is distributed more flexibly across the week and day, whereas \textit{full-time} students follow a more concentrated daytime study pattern. This pattern may reflect structural differences in students’ daily routines, as \textit{part-time} students may more often combine their studies with employment or other daytime obligations and therefore shift learning activities to hours outside standard working times.

\section{Practical Implications}

Overall, our results show that Syntea is already used by a substantial proportion of students, suggesting that AI-based learning has become part of everyday study practice for many learners at IU. At the same time, usage is not evenly distributed across all groups and contexts. Differences by gender, age, study cluster, degree, and study mode indicate that adoption is shaped not only by individual preferences but also by structural and curricular conditions.

The temporal analyses show that Syntea usage is concentrated on weekdays and during daytime to early evening hours, while \textit{part-time} students show relatively stronger activity on weekends and later in the day compared to \textit{full-time} students. These findings have direct practical implications for the design and deployment of AI-based learning support. Platform communication, reminders, and learning nudges should be aligned with empirically observed usage windows rather than standard institutional schedules. In addition, the differences between study modes and age groups suggest that a more adaptive and user-sensitive timing of support features may improve engagement.

%\subsection{Implications for System Development}
From a practical perspective, our findings suggest four priorities for further development. First, coverage should be expanded to course formats in which Syntea is currently unavailable, especially seminars, project-based courses, and thesis-related contexts. Second, more multimodal capabilities may be needed to better support disciplines with visual or non-text-centered learning processes. Third, onboarding and communication strategies should be adapted to different learner groups rather than assuming uniform AI readiness. Fourth, future optimization should focus not only on overall adoption but on contextual fit across programs, study phases, and study modes, and factors such as age or gender. 

\subsection{Limitations}
Our findings should be interpreted in light of several limitations. First, our study is descriptive and does not allow causal conclusions. Second, the analysis is limited to one observation month, which improves comparability but restricts insight into longer-term developments. Third, usage was measured primarily in terms of adoption and temporal patterns, not in terms of interaction quality or direct learning outcomes. Finally, some subgroup results, especially for very small groups, should be interpreted with caution. 

\section{Conclusion and Future Work}
Our study contributes large-scale descriptive evidence on the use of an AI-based learning assistant in higher education and demonstrates that Syntea is already embedded in the learning routines of many distance learners. The results show usage differences across student groups and study contexts. In particular, the findings highlight that adoption is closely linked to course availability, disciplinary fit, and students' everyday study rhythms. However, these patterns should be interpreted in light of structural access conditions and the descriptive nature of the study.

Future work should move beyond descriptive analyses and examine how different patterns of use relate to educational outcomes, such as engagement, persistence, satisfaction, and academic performance. In addition, future studies should include course-level implementation factors in order to distinguish more clearly between learner preferences and structural access conditions. Further research may also investigate interaction intensity and qualitative usage profiles to better understand how different student groups benefit from AI-based learning support. Overall, the present findings provide an empirical basis for the continued refinement of Syntea toward a more adaptive, inclusive, and context-sensitive learning assistant.

\section*{Conflict of Interest}
All authors were employed by IU International University at the time this study was conducted. The authors declare that this affiliation did not influence the study design, analysis, interpretation of the data, or the reporting of results.

\section*{Ethical Impact Statement}
All data were anonymized before analysis. Participation in the learning tool was voluntary and part of a regularly available study support system. The study complied with institutional data protection policies and relevant ethical standards.

%%===========================================================================================%%
%% If you are submitting to one of the Nature Portfolio journals, using the eJP submission   %%
%% system, please include the references within the manuscript file itself. You may do this  %%
%% by copying the reference list from your .bbl file, paste it into the main manuscript .tex %%
%% file, and delete the associated \verb+\bibliography+ commands.                            %%
%%===========================================================================================%%

\bibliography{20-bibliography}% common bib file

%% if required, the content of .bbl file can be included here once bbl is generated
%%\input sn-article.bbl

%% Default %%
%%\input sn-sample-bib.tex%

\end{document}